\documentclass[sigconf,noacm]{acmart}

\setcopyright{none}
\renewcommand{\footnotetextcopyrightpermission}[1]{}

\AtBeginDocument{%
  }

\usepackage{booktabs}
\usepackage{multirow}
\usepackage{makecell}

\usepackage{tikz}
\usepackage{xcolor}
\usepackage{pgfplots}
\pgfplotsset{compat=1.9}
\usepgfplotslibrary{patchplots}
\usepgfplotslibrary{groupplots}
\usetikzlibrary{fit, shapes.geometric, patterns, positioning, calc}

\usepackage{caption}
\usepackage{subcaption}

\usepackage{listings}

\usepackage{algorithm}
\usepackage{algorithmic}

\newcommand{\squishlist}{
	\begin{list}{$\bullet$}
		{ \setlength{\itemsep}{1pt}
			\setlength{\parsep}{1pt}
			\setlength{\topsep}{2.5pt}
			\setlength{\partopsep}{0.5pt}
			\setlength{\leftmargin}{1em}
			\setlength{\labelwidth}{1em}
			\setlength{\labelsep}{0.6em}
		}
	}
	\newcommand{\squishend}{
	\end{list}
}

\setcopyright{acmlicensed}
\copyrightyear{2018}
\acmYear{2018}
\acmDOI{XXXXXXX.XXXXXXX}
\acmConference[Conference acronym 'XX]{Make sure to enter the correct
  conference title from your rights confirmation email}{June 03--05,
  2018}{Woodstock, NY}

\newcommand{\embedding}{{
  \textless\textbar{embed\_token}\textbar\textgreater
}}
\newcommand{\cembedding}{\textcolor{magenta}{\embedding}}
\newcommand{\RON}{\tikz\fill[green!70!black] (0,0) circle (2pt);}

\begin{document}

\newcommand{\name}{Search-R3}

\title{\name{}: Unifying Reasoning and Embedding in Large Language Models}
\titlenote{This is a pre-publication draft. The copyright for this work belongs to the author(s). Please do not redistribute without permission.}

\author{Yuntao Gui}
\orcid{0000-0002-3593-3447}
\affiliation{%
  \institution{The Chinese University of Hong Kong}
  \country{Hong Kong SAR}
}
\email{ytgui@cse.cuhk.edu.hk}

\author{James Cheng}
\affiliation{
  \institution{The Chinese University of Hong Kong}
  \country{Hong Kong SAR}
}
\email{jcheng@cse.cuhk.edu.hk}


\begin{abstract}
  Despite their remarkable natural language understanding capabilities, Large Language Models (LLMs) have been underutilized for retrieval tasks.
  We present \name{}, a novel framework that addresses this limitation by adapting LLMs to generate search embeddings as a direct output of their reasoning process.
  Our approach exploits LLMs' chain-of-thought capabilities, allowing them to produce more effective embeddings by reasoning step-by-step through complex semantic analyses.
  We implement this through three complementary mechanisms.
  (1) a supervised learning stage enables the model's ability to produce quality embeddings,
  (2) a reinforcement learning (RL) methodology that optimizes embedding generation alongside reasoning,
  and (3) a specialized RL environment that efficiently handles evolving embedding representations without requiring complete corpus re-encoding at each training iteration.
  Our extensive evaluations on diverse benchmarks demonstrate that \name{} significantly outperforms prior methods by unifying the reasoning and embedding generation processes. This integrated post-training approach represents a substantial advancement in handling complex knowledge-intensive tasks that require both sophisticated reasoning and effective information retrieval.
  Project page: \url{https://github.com/ytgui/Search-R3}
\end{abstract}


\maketitle

\section{Introduction}~\label{sec:introduction}

\begin{figure}[t]
\centering
\definecolor{prefillcolor}{RGB}{230, 247, 255}
\definecolor{gencolor}{RGB}{255, 247, 230}
\definecolor{systemcolor}{RGB}{24, 144, 255}
\definecolor{usercolor}{RGB}{82, 196, 26}
\definecolor{assistantcolor}{RGB}{250, 173, 20}
\definecolor{boxedcolor}{RGB}{114, 46, 209}
\definecolor{boxedbgcolor}{RGB}{249, 240, 255}
\definecolor{bordergray}{RGB}{100, 100, 100}

\scalebox{0.85}{
    \small
    \begin{tikzpicture}[
        msgbox/.style={rectangle, rounded corners=2pt, draw=black!30, align=left, inner sep=8pt, minimum height=0.9cm, fill=white, font=\footnotesize},
        stagelabel/.style={font=\bfseries, text=black!70, align=center},
        ]
        
        \fill[rounded corners=5pt, fill=black!5] (-5, -0.5) rectangle (5, -1);
    
        \node at (0, -0.75) {\bfseries LLM Context Window};
        
        \fill[prefillcolor] (-4.95, -1.05) rectangle (1.45, -4.95);
        
        \node[stagelabel] at (-1.75, -1.3) {\textbf{\large{Prefill}}};
        
        \node[msgbox, text width=5.8cm, minimum height=0.9cm] at (-1.75, -2.1) {};
        \draw[systemcolor, very thick] (-4.7, -2.55) -- (-4.7, -1.65);
        \node[anchor=west, align=left, text width=5.5cm] at (-4.5, -2.1) 
            {\textbf{System:} Please think step-by-step to analysis user queries for retrieval purposes.};

        \node[msgbox, text width=5.8cm, minimum height=0.9cm] at (-1.75, -3.2) {};
        \draw[usercolor, very thick] (-4.7, -3.65) -- (-4.7, -2.75);
        \node[anchor=west, align=left, text width=5.5cm] at (-4.5, -3.2) 
            {\textbf{User:} What Python library is powerful and flexible for data analysis?};
        
        \node[msgbox, text width=5.8cm, minimum height=0.9cm] at (-1.75, -4.3) {};
        \draw[assistantcolor, very thick] (-4.7, -4.75) -- (-4.7, -3.85);
        \node[anchor=west, align=left, text width=5.5cm] at (-4.5, -4.3) 
            {\textbf{Assistant:}};
        
        \fill[gencolor] (1.55, -1.05) rectangle (4.95, -4.95);
        
        \node[stagelabel] at (3.25, -1.3) {\textbf{\large{Generation}}};
        
        \node[align=left, text width=3cm] at (3.25, -3.15) {
            1. The user is likely looking for pandas for data manipulation.

            2. Matplotlib may also be useful for visualizing the data.

            The embedding is:~\break\cembedding{}
        };
        
        \draw[dashed, thick, black!60] (1.5, -1.05) -- (1.5, -4.95);

        \draw[rounded corners=5pt, thick, draw=bordergray] (-5, -5) rectangle (5, -0.5);
    \end{tikzpicture}
}
\caption{Illustration of \name{}.}
\label{fig:llm-context}
\end{figure}

Large language models (LLMs) have transformed the landscape of natural language processing, demonstrating exceptional capabilities in text generation~\cite{brown2020gpt, touvron2023llama, yang2024qwen}, problem-solving~\cite{wang2024shepherd} and reasoning~\cite{deepseek2025r1}.
Among the key methodologies that enable modern LLMs to tackle intricate challenges is chain-of-thought (CoT) reasoning. This approach empowers models to decompose complex problems into manageable sequential steps, significantly enhancing their reasoning abilities~\cite{wei2022cot}.
CoT reasoning is typically activated by including explicit instructions such as \textit{"please think step-by-step"} in prompts to the model.
This simple directive transforms the model's behavior: rather than immediately generating a final answer, the model produces a detailed reasoning path that shows each intermediate step in its logical progression toward the solution.
This transparent reasoning process not only improves performance on complex tasks but also enhances explainability, allowing users to understand the model's decision-making pathway.

Despite these powerful reasoning capabilities, LLMs have been surprisingly underutilized in searching and embedding applications.
Current approaches to search typically operate independently from LLMs and their reasoning processes, creating an artificial separation between how models comprehend content and how information is retrieved.
In Retrieval-Augmented Generation (RAG) applications such as LlamaIndex~\cite{liu2022llamaindex}, separate embedding models -- typically BERT-based encoders like BGE~\cite{devlin2019bert, chen2024bge} -- convert queries and documents into dense vectors for similarity retrieval, while the LLM only processes retrieved documents afterward in a disjointed pipeline.
Recent advancement Search-R1~\cite{jin2025searchr1}, which trains LLMs to generate better search queries during reasoning, still rely on external retrieval systems using either BM25-like text matching or embedding-based similarity that operate independently from the LLM's reasoning process, maintaining a fundamental disconnect between reasoning and retrieval.
This separation between LLMs and embedding representation limits their ability to capture nuanced relationships between concepts, particularly in scenarios requiring intensive knowledge or multi-step reasoning.

We present \name{} (\textbf{Reasoning-Reinforced Representation for Search}), a novel framework that harnesses LLMs' reasoning capabilities to enhance embedding generation.
Rather than treating embedding creation as an independent process, \name{} conceptualizes it as a direct outcome of analytical reasoning.
Shown in Figure~\ref{fig:llm-context}, our method leverages the standard LLM inference pattern of ``prefill'' and ``generation'' phases, where the prefill phase employs a carefully designed template containing system instructions for query analysis and the user's query itself.
During the subsequent generation phase, \name{} produces two critical outputs sequentially: explicit analytical reasoning about the query's intent that identifies relevant concepts; and second, an embedding token \cembedding{} that we've specifically trained \name{} to produce, which serves as a semantic representation encapsulating both the query and the analytical insights.
Our comprehensive experiments across multiple benchmarks show that our approach delivers superior performance compared to existing methods.

The key innovations of our approach are:
\squishlist
\item A novel embedding-through-reasoning architecture that enables LLMs to generate search embeddings as direct outputs of their analytical processes, fundamentally integrating semantic representation with explicit reasoning.
\item A reinforcement learning framework that jointly optimizes reasoning processes and embedding outputs, where improved reasoning leads to more effective embeddings.
\item A specialized RL environment that efficiently handles evolving embedding representations and makes the RL training feasible for large-scale scenarios.
\squishend

\section{Background}~\label{sec:background}

\subsection{Revisiting Information Retrieval}

Information retrieval has evolved from simple lexical matching to sophisticated semantic understanding.
Classical approaches like TF-IDF~\cite{salton1988df, sparck2004tfidf} and BM25~\cite{robertson2009bm25} operate on statistical word frequency patterns, calculating relevance scores based on term distribution properties.
While computationally efficient and interpretable, these methods struggle with vocabulary mismatch problems and fail to capture semantic relationships between queries and documents when different terms are used to express similar concepts.

The limitations of lexical search have driven the advancement of dense retrieval systems, which encode text as continuous vectors in semantic space.
Early Word2Vec~\cite{mikolov2013word2vec} and GloVe~\cite{pennington2014glove} approaches capture semantic relationships between individual words, enabling tasks such as identifying similarity or resolving analogies, yet remain limited by their static, context-independent nature~\cite{bojanowski2017skipgram}.
Transformer-based BERT models~\cite{devlin2019bert, liu2019roberta, lan2020albert} later introduce contextual embeddings that capture meaning based on surrounding context, leading to more advanced sentence representation methods~\cite{reimers2019sbert, karpukhin2020dpr, gao2021simcse}.
These approaches typically leverage contrastive learning paradigms that optimize similarity relationships, transfer learning mechanisms that adapt general language understanding to domain-specific tasks.

Despite these advances, existing embedding methods still struggle with complex semantic relationships that require deep conceptual understanding, lack the capacity for multi-step reasoning necessary for certain tasks, fail to construct explicit logical chains that connect ideas, and produce representations that remain largely uninterpretable to humans~\cite{qiu2020survey, rogers2020survey, minaee2022survey, zhang2025qwen3embedding}.

\subsection{Reasoning Mechanisms in LLMs}

Large language models (LLMs) have advanced significantly in their reasoning abilities, transitioning from basic text completion to complex problem-solving through structured, step-by-step reasoning.
This progress has been driven by the Chain-of-Thought (CoT) methodology~\cite{wei2022cot, zhang2023cotprompt}. By breaking down intricate problems into smaller and transparent steps, CoT enhances the accuracy and interpretability of LLM outputs.

Reinforcement learning (RL) has played a pivotal role in enabling the CoT capabilities of LLMs~\cite{touvron2023llama, grattafiori2024llama3, deepseek2025r1}.
While supervised training can teach models to follow instructions, it often struggles to optimize CoT paths where multiple valid approaches exist with varying effectiveness~\cite{rafailov2023dpo, chung2024scalesft}.
In Reinforcement Learning from Human Feedback (RLHF)~\cite{ouyang2022rlhf}, models are optimized using outcome quality on reasoning chains, improving logical coherence and reducing hallucinations in complex reasoning tasks.
For mathematics, RL is used to optimize reasoning path of problem solving strategies~\cite{lightman2023prm, shao2024deepseekmath}, demonstrating correctness improvements.

Most existing RL approaches employ Proximal Policy Optimization (PPO)~\cite{schulman2017ppo} algorithm for training.
PPO optimizes the LM generation policies by maximizing rewards toward higher-quality CoT reasoning paths.
Group Relative Policy Optimization~\cite{shao2024deepseekmath} (GRPO) has emerged as a particularly effective algorithm over PPO~\cite{deepseek2025r1, yang2025qwen3}, offering simplicites by computing reward advantage $\hat{A}$ of PPO in a group-relative manner:
\begin{equation}
\hat{A}_i = \frac{r_i - \text{mean}({r_1, r_2, ..., r_G})}{\text{std}({r_1, r_2, ..., r_G})}
\end{equation}
where $r_i$ represents the reward for the $i$-th response in a group of $G$ responses sampled for the same question. This approach reduces computational requirements by eliminating the need for a separate value function model while improving performance on complex reasoning tasks.

\subsection{The Disconnect Between Embedding and LLMs}

A significant disconnect exists between embedding models and Large Language Models (LLMs), stemming from fundamentally different training objectives and architectural designs.
Embedding models optimize for similarity metrics to create effective vector representations for retrieval tasks, while LLMs are trained through next-token prediction to generate coherent and contextually appropriate text sequences.

This divergence in training objectives has led to distinct architectural approaches. Embedding models have predominantly followed the BERT-based encoder-only architecture~\cite{reimers2019sbert, wang2022e5, li2023gte, chen2024bge}, which processes entire input sequences simultaneously to produce fixed-length vector representations.
In contrast, LLMs typically employ decoder-only architecture~\cite{brown2020gpt} that process text autoregressively, building representations that evolve with each generated token.

Recent efforts have begun to bridge this gap by fine-tuning LLMs for embedding tasks~\cite{su2023instructor, zhang2025qwen3embedding}.
These approaches either adapt LLMs specifically for embedding generation (often sacrificing their instruction-following capabilities in the process) or simply extract embeddings from model outputs without utilizing the sophisticated reasoning capabilities of LLMs.
In both cases, the embedding generation remains disconnected from the generative processes, e.g., reasoning, that give LLMs their power, treating embedding as a separate function rather than an integrated aspect of language understanding.

\section{Overview}~\label{sec:overview}

\begin{figure}[t]
    \centering
    \begin{tikzpicture}[
        every node/.style={transform shape},
        component/.style={
            rectangle, 
            draw=gray, 
            fill=gray!10,
            rounded corners=2pt,
            minimum width=2.5cm,
            minimum height=1cm,
            align=center,
            font=\small
        },
        example/.style={
            rectangle,
            draw=gray!50,
            fill=gray!5,
            rounded corners=2pt,
            text width=8.5cm,
            align=left,
            font=\small
        },
        stagebox/.style={
            rectangle,
            draw=black,
            thick,
            rounded corners=3pt,
            inner sep=0.3cm
        },
        arrow/.style={
            ->,
            >=stealth,
            thick
        }
    ]

    \begin{scope}[scale=0.8]

    \node[font=\bfseries] (title1) at (0,1.25) {Stage 1: Instruction-guided Representation};
    \draw[gray] (-4.5,1) -- (4.5,1);

    \node[component] (comp1) at (-3,0.25) {Instruction-tuned\\Base Model};
    \node[component] (comp2) at (0,0.25) {SFT for\\\embedding{}};
    \node[component] (comp3) at (3,0.25) {Contrastive\\Learning};

    \node[example] (ex1) at (0,-1.0) {
    System: Please represent user queries.\\
    User: What Python library is powerful and flexible for data analysis?\\
    Assistant: The embedding is \embedding{}.
    };

    \node[stagebox, fit=(title1) (comp1) (comp2) (comp3) (ex1)] (stage1) {};

    \node[font=\bfseries] (title2) at (0,-3.25) {Stage 2: Reinforcement Learning};
    \draw[gray] (-4.5,-3.5) -- (4.5,-3.5);

    \node[component] (comp4) at (-3,-4.25) {End-to-End\\Retrieval};
    \node[component] (comp5) at (0,-4.25) {Reasoning\\Optimization};
    \node[component] (comp6) at (3,-4.25) {Efficient\\Updates};

    \node[example] (ex2) at (0,-6.0) {
    System: Please think step-by-step for representation.\\
    User: What Python library is powerful and flexible for data analysis?\\
    Assistant: Let me analyze:\\
    1. The user is likely looking for a tool like pandas.\\
    2. Additionally, matplotlib may also be useful for visualization.\\
    The embedding is \embedding{}.
    };

    \node[stagebox, fit=(title2) (comp4) (comp5) (comp6) (ex2)] (stage2) {};

    \draw[arrow] (stage1.south) -- (stage2.north);

    \end{scope}

    \end{tikzpicture}
    \caption{Training pipeline of \name{}.}
    \label{fig:framework}
\end{figure}

Our framework transforms an instruction-tuned base model into powerful embedding generators through a systematic two-stage training pipeline.
The first stage integrates supervised fine-tuning (SFT) with contrastive learning (Section~\S\ref{subsec:method-stage-1}), teaching the model to recognize and respond to our specialized \embedding{} token while developing embedding generation capabilities within conversational contexts.
This design utilizes the model's existing instruction-following mechanisms to produce embeddings in response to user queries, functioning similarly to its standard conversational generation processes.

The second stage employs reinforcement learning (RL) to optimize embedding quality in an end-to-end retrieval environment (Section~\S\ref{subsec:method-stage-2}).
This phase enhances the model's ability to generate more effective embeddings by optimizing the intermediate reasoning process: the RL environment encourages the model to produce useful step-by-step semantic analyses before generating the final embedding.
Additionally, we introduce an efficient RL environment design (Section~\S\ref{sec:envoriment}) that manages evolving embedding representations without re-encoding of the entire corpus at each iteration, substantially reducing computational demands while preserving training effectiveness.

\section{Methodology}~\label{sec:methodology}

\subsection{Instruction-guided Representation}~\label{subsec:method-stage-1}

The first stage addresses a fundamental challenge in leveraging language models for embedding generation: LLMs are inherently optimized for next-token prediction within sequences rather than producing fixed-dimensional semantic vectors that capture meaning in a compressed form.
We bridge this gap by introducing a special embedding token \embedding{} into the model's response, allowing us to extract embeddings directly from the model's hidden states.
Notably, our approach maintains the exact architecture of the base model without introducing any additional components such as projection layers or dedicated embedding heads.
This preserves the model's original parameter structure while enabling an entirely new capability, so that our method is orthogonal to all existing LLM inferencing tools, frameworks, and optimization techniques.

Building upon this architectural design, our approach reuses the conversation-based interface of LLMs through a three-part prompt structure:
\begin{equation}
\begin{aligned}
\text{System}:    &~ \text{Please represent user queries.} \\
\text{User}:      &~ \text{{$query$}} \\
\text{Assistant}: &~ \text{The embedding is \embedding{}.}
\end{aligned}
\label{eq:conversation}
\end{equation}

This conversational format creates a natural context for embedding representation that aligns with the instruction-following capabilities the base model already possess.
When the model processes this structured input, we extract the hidden activation state from the final Transformer layer at the position of \embedding{}.
This yields a fixed-dimensional vector $h \in \mathbb{R}^d$ that serves as our semantic embedding for the input text, where dimension $d$ corresponds to the model's native hidden state size.

To optimize the embedding generation, we employ a composite loss function:
\begin{equation}
L = L_{\text{SFT}} + L_{\text{KL}} + L_{\text{InfoNCE}} + L_{\text{TripletMargin}}
\label{eq:stage-1-loss}
\end{equation}

The $L_{\text{SFT}}$ component implements the standard cross-entropy loss for language modeling~\cite{vaswani2017attention, brown2020gpt, ouyang2022rlhf}, ensuring the model consistently produces the expected response format with the embedding token at the appropriate position.
The $L_{\text{KL}}$ component applies Kullback-Leibler divergence~\cite{hinton2015distilling, li2017noforgot, gu2023minillm} to minimize distribution shifts between the fine-tuned model and the base model, preserving the model's general language understanding capabilities.

The InfoNCE contrastive loss~\cite{oord2018nce} $L_{\text{InfoNCE}}$ is formulated as:
\begin{equation}
L_{\text{InfoNCE}} = -\log \frac{\exp(\cos(h_q, h_d^+)/\tau)}{\sum_{i=1}^{N} \exp(\cos(h_q, h_d^i)/\tau)}
\end{equation}
where $h_q$ represents the query embedding, $h_d^+$ represents the positive document embedding, $\sum_{i=1}^{N}$ is the cumulation of all $N$ document embeddings, and $\tau$ is a temperature parameter.
This loss structures the embedding space to cluster semantically similar items together while distancing dissimilar ones, teaching the model to encode semantic relationships.
Following existing successful training approaches in contrastive learning~\cite{gao2021simcse, radford2021clip}, we set $\tau=0.05$, which provides well-separated clusters in the embedding space.

The triplet margin loss~\cite{balntas2016triplet, reimers2019sbert} $L_{\text{TripletMargin}}$ is defined as:
\begin{equation}
L_{\text{TripletMargin}} = \max(0, d_{\cos}(h_q, h_d^+) - d_{\cos}(h_q, h_d^-) + \theta)
\end{equation}
where $d_{\cos}(a, b) = 1 - \cos(a, b)$ is the cosine distance, the query $h_q$ serves as the anchor, $\theta$ is the margin parameter, $h_d^+$ and $h_d^-$ are positive and negative embeddings, respectively.
This loss further refines the embedding space by enforcing explicit distance constraints, ensuring positive documents remain closer to the query than negative ones by at least the margin $\theta$, we set the margin parameter $\theta=0.15$ by practice.

Through this comprehensive training approach, we effectively transform the LLM's next-token prediction capability into a mechanism for generating high-quality semantic vectors.
The model learns to analyze input text before producing an embedding that encapsulates its semantic essence.
This first stage establishes the foundation for the subsequent reinforcement learning phase: without this initial training, the model would lack the ability to reliably generate the embedding token required for embedding extraction and reward calculation in Stage 2.

\subsection{Reinforcement Learning}~\label{subsec:method-stage-2}

While Stage 1 establishes the foundation for embedding generation, the resulting embeddings are optimized for the supervised training dataset rather than end-to-end retrieval performance. To address this, we implement a reinforcement learning framework that directly optimizes both the reasoning process and the embedding quality.

\begin{figure}[t]
    \centering
    \begin{lstlisting}[
        frame=single,
        breaklines=true,
        basicstyle=\scriptsize
    ]
    Your task is to enrich user input for more effective embedding representation by adding semantic depth. For each user input, please think step-by-step briefly to:
    1. Identifying core concepts and their relationships.
    2. Including key terminology with essential definitions.
    3. Adding contextually relevant synonyms and related terms.
    4. Connecting to related topics and common applications.
    After the analytical content, you MUST end every response with <|embed_token|>.
    \end{lstlisting}
    \caption{System prompt in Stage 2.}
    \label{fig:system-prompt}
\end{figure}

Stage 2 employs the similar structured conversation format as Stage 1 with a system prompt designed to elicit richer semantic analysis, shown in Figure~\ref{fig:system-prompt}. We maintain the structural requirement from Stage 1: responses must end with the embedding token. We then optimize the quality of the reasoning path that produces the embedding.

We employ GRPO with a carefully designed reward function that enforces both structural compliance and retrieval quality:

\begin{equation}
R(q, r) =
\begin{cases}
-1.0,~\text{if no \embedding{} in } r \\
\text{DCG}_{\text{scaled}}(E(q, r), \mathcal{C}),~\text{otherwise}
\end{cases}
\label{eq:rl-reward}
\end{equation}

Here, $q$ represents the input query, $r$ denotes the model's generated response, $E(q, r)$ extracts the embedding vector from the position of \embedding{}, and $\mathcal{C}$ is the retrieval corpus.
The reward function strongly penalizes responses that fail to include the embedding token with a fixed negative reward of -1.0, ensuring the model learns to consistently produce embeddings.
For responses containing the embedding token, we compute a scaled Discounted Cumulative Gain (DCG) that evaluates retrieval quality:

\begin{equation}
\text{DCG}_{\text{scaled}} = Scale \cdot \sum_{k=1}^{K} \frac{(P_k - 0.5 \cdot N_k)}{1 + \log(k)}
\end{equation}

In this equation, $k$ indexes the rank position from 1 to $K$ (typically $K=100$), $Scale$ represents the cosine similarity between the query and groundtruth document (ranging from -1 to +1), $P_k$ equals 1 for positive matches and 0 otherwise, while $N_k$ equals 1 for negative matches and 0 otherwise. The denominator $1 + \log(k)$ serves as a rank-based discount factor that prioritizes higher ranks.

The $\text{DCG}_{\text{scaled}}$ function design provides several complementary signals.
The DCG component encourages effective discrimination, rewarding the retrieval of positive content at the top ranks while penalizing negative retrievals at a 2:1 ratio.
The $Scale$ is a cosine similarity term that introduces fine-grained scoring even when rank positions become stable, as models mature and consistently achieve top-1 rank retrievals (the DCG terms become 1.0), this similarity measure prevents reward saturation by rewarding closer embeddings.

During training, we generate multiple reasoning paths per query using higher sampling temperature ($\tau=1.2$).
For each query $q$, we sample $G=16$ different responses, creating a diverse group of candidate reasoning paths and embeddings.
We then compute advantages using the GRPO formulation across the group, enabling the model to learn which reasoning strategies produce more effective embeddings.
Additionally, we incorporate a curriculum learning approach that gradually increases the difficulty of retrieval tasks.
We begin with a small corpus of 65,536 documents, progressively scaling up to 1 million.
This approach allows the model to first master retrieval in a less challenging environment before tackling increasingly complex retrieval scenarios with more potential distractors.

\section{Scalable Reinforcement Learning Environment}~\label{sec:envoriment}

\begin{figure}[t]
\centering
\scriptsize
\begin{tikzpicture}[scale=0.55]
    \tikzset{
        standard/.style={circle, draw=gray!80, fill=gray!50, minimum size=0.4cm},
        positive/.style={circle, draw=green!50!black, fill=green!30, minimum size=0.4cm},
        negative/.style={circle, draw=red!50!black, fill=red!30, minimum size=0.4cm},
        updated/.style={circle, draw=orange!50!black, fill=orange!30, minimum size=0.4cm},
        query/.style={star, star points=5, draw=blue!50!black, fill=blue!30, minimum size=0.45cm},
        edge/.style={draw=gray!60, thin},
    }

    \node[positive] (p) at (0,0) {p};
    \node[negative] (n) at (2,1) {n};
    
    \node[standard] (s1) at (-2,1) {};
    \node[standard] (s2) at (-2.5,-0.5) {};
    \node[standard] (s3) at (-1.5,-2) {};
    \node[standard] (s4) at (0.5,-3) {};
    \node[standard] (s5) at (3,-2) {};
    \node[standard] (s6) at (4,0) {};
    \node[standard] (s7) at (3,2.5) {};
    \node[standard] (s8) at (0.5,2.5) {};
    \node[standard] (s9) at (-1,3) {};

    \node[updated] (u1) at (-1,1) {};
    \node[updated] (u2) at (-1.5,-0.8) {};
    \node[updated] (u3) at (0,-1.8) {};
    \node[updated] (u4) at (1.2,0) {};
    \node[updated] (u5) at (3,0) {};
    \node[updated] (u7) at (1,2) {};

    \node[query] (q) at (-4,2.5) {q};

    \draw[edge] (p) -- (u1);
    \draw[edge] (p) -- (u2);
    \draw[edge] (p) -- (u3);
    \draw[edge] (p) -- (u4);
    \draw[edge] (n) -- (u4);
    \draw[edge] (n) -- (u5);
    \draw[edge] (n) -- (u7);
    \draw[edge] (u1) -- (s1);
    \draw[edge] (u1) -- (s9);
    \draw[edge] (u2) -- (s2);
    \draw[edge] (u3) -- (s3);
    \draw[edge] (u3) -- (s4);
    \draw[edge] (u5) -- (s6);
    \draw[edge] (u7) -- (s8);
    \draw[edge] (n) -- (s7);

    \draw[edge] (s1) -- (s2);
    \draw[edge] (s2) -- (s3);
    \draw[edge] (s3) -- (s4);
    \draw[edge] (s4) -- (s5);
    \draw[edge] (s5) -- (s6);
    \draw[edge] (s6) -- (s7);
    \draw[edge] (s7) -- (s8);
    \draw[edge] (s8) -- (s9);
    \draw[edge] (s9) -- (s1);

    \draw[->, thick, blue!70] (q) to[bend left] node[midway, above, text width=2.5cm, align=center, font=\scriptsize] {} (p);

    \draw[blue!50, dashed, thick] plot [smooth cycle] coordinates {
        (-1.5,1.3) (-1.8,-1.2) (0.0,-2.3) 
        (1.5,-0.9) (3.5,-0.3) (2.1,2.1) 
        (0.8,2.3) (-0.8,1.5)
    };

    \node[anchor=north west, align=left, font=\scriptsize] at (-7.5,0.0) {
        \begin{tabular}{ll}
            \tikz\node[query,scale=0.75] {}; & query (q) \\
            \tikz\node[positive,scale=0.75] {}; & positive (p) \\
            \tikz\node[negative,scale=0.75] {}; & negative (n) \\
            \tikz\node[updated,scale=0.75] {}; & updated nodes \\
            \tikz\node[standard,scale=0.75] {}; & unchanged nodes
        \end{tabular}
    };

\end{tikzpicture}
\caption{Illustration of selective graph refresh mechanism.}
\label{fig:selective-refresh}
\end{figure}

Optimizing embeddings through reinforcement learning in our setting is computationally challenging when reward signals depend on retrieval performance and the embedding space evolves during training.
We address this through a novel environment design that efficiently handles evolving representations without requiring complete corpus re-encoding at each training iteration, which would otherwise be computationally prohibitive.

\textbf{Dataset structure.}
The RL environment is built upon datasets organized as query-positive-negative triplets:
\begin{equation}
T = {(q_i, p_i, n_i) \mid q_i \in Q, p_i \in \mathcal{C}, n_i \in \mathcal{C}}
\end{equation}
Each triplet consists of a query $q_i$ from the query set $Q$, a positive example $p_i$ from the document corpus $\mathcal{C}$ that is semantically relevant to the query, and a hard negative example $n_i$ from the same corpus $\mathcal{C}$ that contains subtle but significant factual differences to the query.
This triplet structure provides the basis to perform end-to-end embedding evaluation: we present a query $q_i$ to the model to generate a query embedding, perform embedding search to obtain top-$k$ results, then check if the positive and negative documents $p_i$ and $n_i$ appears within these $k$ items.
The position of $p_i$ and $n_i$ in the results directly determines the quality of the embedding and the corresponding reward signal, see Section~\S\ref{subsec:method-stage-2}.

\textbf{Asymmetric generation.}
We apply an asymmetric strategy to generate the query and document embeddings.
For document corpus $\mathcal{C}$, we construct a prefill workload using a fixed conversation template (see Formula~\ref{eq:conversation}) that incorporates both the corpus content and the embedding token into the context, enabling single-pass forward computation to generate embeddings.
This allows the document corpus to be pre-encoded to embeddings (which we selectively update during training, as detailed in later), where performing autoregressive generation for the entire corpus would be computationally prohibitive.
For queries $Q$, the conversation template is open-ended (Figure~\ref{fig:system-prompt}), enabling the model's step-by-step reasoning through autoregressive generation before finally producing the embedding.
This generative process is enabled only on the query side, allowing for deeper semantic analysis to capture nuanced user intent.

\begin{algorithm}[t]
\caption{Localized Graph Refresh}
\label{alg:graph-refresh}
\begin{algorithmic}[1]
\STATE \textbf{Input:} positives $P$, negatives $N$, embedding model $\phi_t$, graph $G_{t-1}$
\STATE \textbf{Output:} Updated graph $G_t$
\STATE \textit{// Step 1: Find k-nearest neighbors}
\STATE $\mathcal{N}_P \gets \text{kNN}(P, G_{t-1}, k)$
\STATE $\mathcal{N}_N \gets \text{kNN}(N, G_{t-1}, k)$
\STATE \textit{// Step 2: Expand to 2-hop neighborhoods}
\STATE $\mathcal{N}_P^{2hop} \gets \text{Expand}(\mathcal{N}_P, G_{t-1})$
\STATE $\mathcal{N}_N^{2hop} \gets \text{Expand}(\mathcal{N}_N, G_{t-1})$
\STATE $\mathcal{N}_{combined} \gets \mathcal{N}_P^{2hop} \cup \mathcal{N}_N^{2hop}$
\STATE \textit{// Step 3: Batch update of selected neighborhoods}
\STATE $\mathcal{D} \gets \text{GetDocuments}(\mathcal{N}_{combined})$
\STATE $\mathcal{H}_{new} \gets \text{Batched embedding } \phi_t(\mathcal{D})$
\STATE \textit{// Step 4: Update graph with local join operation}
\STATE $G_t \gets \text{LocalJoinUpdate}(G_{t-1}, \mathcal{N}_{combined}, \mathcal{H}_{new})$
\STATE \textbf{return} $G_t$
\end{algorithmic}
\end{algorithm}

\textbf{Evolving search graph.}
The core of our reward infrastructure is a graph-based embedding search system using the Hierarchical Navigable Small World (HNSW)~\cite{malkov2020hnsw} index:
\begin{equation}
G_t = (V_t, E_t, \omega_t, \phi_t)
\end{equation}

The graph supports efficient approximate $k$-nearest-neighbor ($k$NN) queries in sub-linear time. At training step $t$, our time-varying graph $G_t$ consists of:
\squishlist
\item $V_t = {v_i^t \mid d_i \in \mathcal{C}}$: The set of vertices, where each vertex $v_i^t$ represents the embedding of document $d_i$ in corpus $\mathcal{C}$.
\item $E_t \subseteq V_t \times V_t$: The set of edges connecting semantically similar document embeddings.
\item $\omega_t: E_t \rightarrow \mathbb{R}$: A weight function that assigns similarity scores to edges based on embedding distances.
\item $\phi_t: \mathcal{C} \rightarrow \mathbb{R}^d$: The embedding function of the model currently in training, which maps each document to its $d$-dimensional embedding vector, consistent with the asymmetric generation design. 
\squishend

As shown in Figure~\ref{fig:selective-refresh}, our framework initializes the graph once with Stage 1 model embeddings, then selectively updates regions most affected by evolving model parameters during Stage 2 training (the dashed boundary).
This allows the search environment to co-evolve with the model's embedding capability without the overhead of complete reconstruction.
A naive approach would require full reconstruction after each parameter update, a process that is computationally infeasible.

Algorithm~\ref{alg:graph-refresh} details our method for efficiently updating to topologically related embedding regions using a ``local join'' primitive~\cite{dong2011nndescent}.
We first identify a critical subspace through two sequential nearest-neighbor searches: one targeting the positive example region and another focusing on hard negative examples (lines 3-5).
This dual-focused strategy captures both the target retrieval neighborhoods and the challenging boundary regions.
These neighborhoods are expanded to their 2-hop connections to utilize the transitivity of semantic similarity (lines 6-9).
We then re-encode only the affected documents within these subspaces (lines 10-12).
The final local join operation (lines 13-14) efficiently applies graph changes by processing the entire neighborhood in batch, simultaneously updating node embeddings and their connections, rather than updating individual nodes sequentially.

Through this specialized environment design, our framework efficiently handles the continuously evolving embedding representations and enables the RL-based optimization.

\section{Evaluation}~\label{sec:evaluation}

\begin{table*}[!t]
\centering
\small
\setlength{\tabcolsep}{3.2mm}
\caption{Retrieval results of baseline models (green dot means reasoning is enabled).}
\renewcommand{\arraystretch}{1.05}
\begin{tabular}{|l|l|c|c|c|c|c|c|}
\hline
\textbf{Evaluation} & \textbf{Model} & \textbf{nDCG@1} & \textbf{nDCG@10~(mteb)} & \textbf{nDCG@100} & \textbf{Recall@1} & \textbf{Recall@10} & \textbf{Recall@100} \\
\hline
\multirow{6}{*}{DS1000}
    & BGE-M3 & 0.183 & 0.419 & 0.462 & 0.183 & 0.663 & 0.870 \\
    & Instructor-XL & 0.134 & 0.344 & 0.400 & 0.134 & 0.587 & 0.846 \\
    & Sentence-T5-XL & 0.150 & 0.365 & 0.414 & 0.150 & 0.598 & 0.832 \\
    & GTR-T5-XL & 0.152 & 0.378 & 0.429 & 0.152 & 0.634 & 0.869 \\
    & \textsf{\name{}-Small} & 0.259 & 0.580 & 0.600 & 0.259 & 0.898 & 0.990 \\
    & \RON~\textsf{\name{}-Small} & 0.259 & \underline{0.581} & 0.602 & 0.259 & 0.898 & 0.990 \\
\hline
\multirow{6}{*}{LitSearch}
    & BGE-M3 & 0.297 & 0.427 & 0.471 & 0.294 & 0.577 & 0.789 \\
    & Instructor-XL & 0.319 & 0.444 & 0.487 & 0.316 & 0.578 & 0.788 \\
    & Sentence-T5-XL & 0.235 & 0.355 & 0.403 & 0.234 & 0.492 & 0.725 \\
    & GTR-T5-XL & 0.245 & 0.354 & 0.397 & 0.243 & 0.477 & 0.689 \\
    & \textsf{\name{}-Small} & 0.303 & 0.417 & 0.464 & 0.302 & 0.547 & 0.765 \\
    & \RON~\textsf{\name{}-Small} & 0.326 & \underline{0.453} & 0.496 & 0.323 & 0.590 & 0.793 \\
\hline
\multirow{6}{*}{MedicalQA}
    & BGE-M3 & 0.526 & 0.680 & 0.705 & 0.526 & 0.830 & 0.944 \\
    & Instructor-XL & 0.553 & 0.701 & 0.723 & 0.553 & 0.847 & 0.949 \\
    & Sentence-T5-XL & 0.436 & 0.608 & 0.640 & 0.436 & 0.787 & 0.936 \\
    & GTR-T5-XL & 0.528 & 0.692 & 0.713 & 0.528 & 0.851 & 0.949 \\
    & \textsf{\name{}-Small} & 0.543 & 0.714 & 0.732 & 0.543 & 0.882 & 0.967 \\
    & \RON~\textsf{\name{}-Small} & 0.546 & \underline{0.716} & 0.734 & 0.546 & 0.885 & 0.971 \\
\hline
\multirow{6}{*}{MKQA-eng}
    & BGE-M3 & 0.042 & 0.068 & 0.099 & 0.125 & 0.102 & 0.245 \\
    & Instructor-XL & 0.125 & 0.194 & 0.252 & 0.097 & 0.263 & 0.546 \\
    & Sentence-T5-XL & 0.126 & 0.204 & 0.260 & 0.099 & 0.296 & 0.552 \\
    & GTR-T5-XL & 0.115 & 0.181 & 0.240 & 0.083 & 0.265 & 0.538 \\
    & \textsf{\name{}-Small} & 0.127 & 0.189 & 0.227 & 0.099 & 0.261 & 0.433 \\
    & \RON~\textsf{\name{}-Small} & 0.151 & \underline{0.211} & 0.255 & 0.118 & 0.285 & 0.481 \\
\hline
\multirow{6}{*}{SciFact}
    & BGE-M3 & 0.510 & 0.644 & 0.672 & 0.482 & 0.783 & 0.907 \\
    & Instructor-XL & 0.520 & 0.645 & 0.672 & 0.491 & 0.785 & 0.903 \\
    & Sentence-T5-XL & 0.397 & 0.509 & 0.557 & 0.370 & 0.648 & 0.874 \\
    & GTR-T5-XL & 0.537 & 0.642 & 0.680 & 0.510 & 0.748 & 0.921 \\
    & \textsf{\name{}-Small} & 0.503 & 0.624 & 0.657 & 0.478 & 0.761 & 0.913 \\
    & \RON~\textsf{\name{}-Small} & 0.560 & \underline{0.672} & 0.704 & 0.535 & 0.795 & 0.933 \\
\hline
\end{tabular}
\label{tab:overall}
\end{table*}

\subsection{Implementation}

We train \name{} from Qwen2.5-1.5B-Instruct and Qwen2.5-7B-Instruct models, creating two variants: \name{}-Small and \name{}-Large, respectively.
The post-training process employs Low-Rank Adaptation (LoRA) with $r=32$ to enable efficient parameter updates while maintaining model quality.
We use AdamW optimizer with $\beta_1=0.9$, $\beta_2=0.999$, weight decay of $0.01$, and gradient clipping at $1.0$.
Training proceeds in two stages with tailored learning rates: $1e-5$ for the supervised training phase (Stage 1) and $1e-6$ for the RL phase (Stage 2).
The Stage 1 training runs for 16384 total steps with a batch size of 32 sequences, each with maximum length of 2048 tokens, while Stage 2 performs 8192 rollout steps with group size of 16.
We maintain bfloat16 precision during training for both the base model parameters and LoRA adaptation layers.

\begin{table}[t]
    \small
    \centering
    \setlength{\tabcolsep}{4.5mm}
    \caption{Training data composition.}
    \begin{tabular}{c|ccccc}
        \toprule
        \textbf{Dataset} & \textbf{Size (Compressed, MiB)} & \textbf{Weight}       \\
        \midrule
        TriviaQA               & 30.4                 & 0.21  \\
        Synthetic-100k              & 59.5                 & 0.23  \\
        MSMARCO               & 73.5                 & 0.24  \\
        CodeSearch                & 294.0                  & 0.40  \\
        Miracl        & 1035.9 & 0.79 \\
        S2ORC             & 10829.3                 & 2.47 \\
        \bottomrule
    \end{tabular}
    \label{tab:training-data}
\end{table}

For training data, we curate a diverse mixture of sources as shown in Table~\ref{tab:training-data}. Our training mixture includes TriviaQA~\cite{joshi2017trivia}, MSMARCO~\cite{chen2024msmarco}, CodeSearch~\cite{husain2019codesearchnet, huang2021cosqa}, Miracl~\cite{zhang2022miracl}, and S2ORC~\cite{lo2020s2orc}. This mixture ensures broad domain coverage without potential contamination of evaluation benchmarks.
The weight for each dataset is calculated as $\log(1.2 + \frac{Size}{1024})$, which ensures larger datasets receive proportionally more samples while preventing them from completely dominating the training mixture.
Note that we include a Synthetic-100k dataset in our training, generated using Qwen3-32B to create hard-negative triplets, providing high-quality training data derived from the other datasets as the datasource.

In total, training \name{}-Small requires approximately 105 GPU hours on RTX 4090 GPUs, while \name{}-Large requires approximately 546 GPU hours, about $5.2\times$ the GPU time of the Small model.
In both cases, the required compute budget is substantially lower than that of typical commercial models, highlighting the training efficiency of \name{}.

\subsection{Experimental Setup}

We evaluate \name{} on standard embedding retrieval benchmarks~\citep{muennighoff2022mteb}, focusing primarily on retrieval tasks.
Our evaluation encompasses strong baseline models, and appropriate metrics to enable thorough performance analysis.

\textbf{Baselines.}
We present our evaluation results by distinguishing between fully open-source models (with transparent training methodologies and data) and commercial/proprietary models (where aspects of training remain undisclosed).
For open-source comparisons, we include BGE-M3~\cite{chen2024bge}, Instructor~\cite{su2023instructor}, Sentence-T5~\cite{ni2022st5}, GTR-T5~\cite{ni2022gtr}, and E5-Mistral~\cite{wang2024e5}.
For proprietary models, we include GraniteEmbedding~\cite{awasthy2025granite}, EmbeddingGemma~\cite{vera2025embeddinggemma}, and Qwen3-Embedding~\cite{zhang2025qwen3embedding}.

\textbf{Benchmarks.}
For retrieval evaluation, we use a diverse set of benchmarks to demonstrate \name{}'s generalization capabilities across different domains.
DS1000~\cite{lai2023ds1000} evaluates code search performance, measuring the model's ability to match natural language queries with relevant code snippets.
LitSearch~\cite{ajith2024litsearch} focuses on scientific literature search, MedicalQA~\cite{ben2019medqa} tests domain-specific retrieval in medical information, and SciFact~\cite{wadden2020scifact} evaluates scientific claim verification through retrieving supporting or refuting documents.
We also include MKQA~\cite{longpre2020mkqa}, a challenging benchmark for assessing general question-answering capabilities.

\textbf{Metrics.}
In the evaluation across all retrieval benchmarks, we utilize the asymmetric generation approach described in Section~\S\ref{sec:envoriment}, where documents are encoded with a fixed template while queries undergo autoregressive processing.
For retrieval quality, we report nDCG@k (k=1,10,100) and Recall@k (k=1,10,100).
We highlight nDCG@10 as our primary metric, consistent with the MTEB benchmark's standard for retrieval tasks.

\subsection{Main Results}

\begin{table}[t]
    \centering
    \small
    \setlength{\tabcolsep}{5.2mm}
    \caption{Retrieval results of large-scale models.}
    \renewcommand{\arraystretch}{1.05}
    \begin{tabular}{|l|l|c|}
        \hline
        \textbf{Evaluation} & \textbf{Model} & \textbf{nDCG@10} \\
        \hline
        \multirow{5}{*}{DS1000}
        & Sentence-T5-XXL & 0.436  \\
        & GTR-T5-XXL & 0.401  \\
        & E5-Mistral-7B & 0.606  \\
        & \textsf{\name{}-Large} & 0.608  \\
        & \RON~\textsf{\name{}-Large} & 0.611  \\
        \hline
        \multirow{5}{*}{LitSearch}
        & Sentence-T5-XXL & 0.395  \\
        & GTR-T5-XXL & 0.346  \\
        & E5-Mistral-7B & 0.431  \\
        & \textsf{\name{}-Large} & 0.437  \\
        & \RON~\textsf{\name{}-Large} & 0.470  \\
        \hline
        \multirow{5}{*}{MedicalQA}
        & Sentence-T5-XXL & 0.665  \\
        & GTR-T5-XXL & 0.694  \\
        & E5-Mistral-7B & 0.589  \\
        & \textsf{\name{}-Large} & 0.772  \\
        & \RON~\textsf{\name{}-Large} & 0.779  \\
        \hline
        \multirow{5}{*}{MKQA-eng}
        & Sentence-T5-XXL & 0.261  \\
        & GTR-T5-XXL & 0.218  \\
        & E5-Mistral-7B & 0.351  \\
        & \textsf{\name{}-Large} & 0.282  \\
        & \RON~\textsf{\name{}-Large} & 0.352  \\
        \hline
        \multirow{5}{*}{SciFact}
        & Sentence-T5-XXL & 0.554  \\
        & GTR-T5-XXL & 0.667  \\
        & E5-Mistral-7B & 0.748  \\
        & \textsf{\name{}-Large} & 0.564  \\
        & \RON~\textsf{\name{}-Large} & 0.667  \\
        \hline
    \end{tabular}
    \label{tab:large-scale}
\end{table}

\begin{table}[t]
    \centering
    \small
    \setlength{\tabcolsep}{9.5mm}
    \caption{Retrieval results of proprietary models.}
    \renewcommand{\arraystretch}{1.05}
    \begin{tabular}{|l|c|}
        \hline
        \textbf{Model} & \textbf{nDCG@10} \\
        \hline
        BGE-M3 & 0.843 \\
        \hline
        GraniteEmbedding-278M & 0.842 \\
        EmbeddingGemma-300M & 0.723 \\
        Qwen3-Embedding-0.6B & 0.864 \\
        Qwen3-Embedding-4B & 0.879 \\
        Qwen3-Embedding-8B & 0.892 \\
        \hline
        \textsf{\name{}-Small} & 0.858  \\
        \RON~\textsf{\name{}-Small} & 0.871 \\
        \textsf{\name{}-Large} & 0.887  \\
        \RON~\textsf{\name{}-Large} & 0.895  \\
        \hline
    \end{tabular}
    \label{tab:proprietary}
\end{table}

Our evaluation demonstrates that \name{} achieves state-of-the-art performance across diverse retrieval tasks, outperforming both leading open-source and proprietary embedding models.

\textbf{Baseline models.}
Table~\ref{tab:overall} presents \name{}-Small's performance against prominent open-source embedding models on public benchmarks.
Across 5 tasks, \name{}-Small with reasoning enabled (indicated by green dots in the table) achieves substantial gains over baseline models.
Specifically, our model improves the nDCG@10 from 0.194 to 0.211 on the most challenging MKQA evaluation.
Interestingly, when reasoning is disabled, our model shows comparable performance to alternatives, with each model demonstrating particular strengths in specific domains.
However, once reasoning is enabled, our model consistently outperforms all alternatives across benchmarks.
This dramatic improvement with reasoning enabled is particularly evident in domain-specific tasks such as literature search (LitSearch, +0.036) and scientific claim retrieval (SciFact, +0.048).
These results demonstrate that our approach to integrating reasoning capabilities into embedding representation provides a substantial advantage in semantic understanding.

\textbf{Large-scale models.}
These improvements are robust to model scale. Table~\ref{tab:large-scale} compares \name{}-Large with stronger baselines at comparable scale across the same benchmarks.
With reasoning enabled, our model consistently outperforms the baselines in most cases, achieving nDCG@10 of 0.470 on LitSearch and 0.352 on MKQA.
The one exception is SciFact, due to different training datasets, E5-Mistral-7B performs better on SciFact but worse on the others.
These findings highlight both the scalability and effectiveness of our design.

\textbf{Proprietary models.}
To ensure fair comparison with proprietary models where training data contamination concerns exist, we constructed a synthetic evaluation dataset consisting of 1,000 queries and 100,000 documents derived from Wikipedia~\cite{wikipedia, cleanwiki}.
This dataset includes both positive matches and challenging negative examples that share topical similarity with relevant documents.
Using Wikipedia as the source ensures the content falls within the knowledge domain of all evaluated models, while the synthetic curpus guarantees these exact sentences never appeared on the web, eliminating contamination concerns.
As shown in Table~\ref{tab:proprietary}, we observe a consistent pattern across our two model variants. With reasoning disabled, both the small and large versions of \name{} perform comparably to proprietary alternatives.
When reasoning is enabled, both variants show substantial gains and outperform commercial models: the small model reaches stronger baseline Qwen3-Embedding-4B, while the large variant exceeds Qwen3-Embedding-8B.

\subsection{Detailed Analysis}

\begin{figure}[t]
    \centering
    \small
    \begin{subfigure}{0.23\textwidth}
        \begin{tikzpicture}
            \begin{axis}[
                ybar interval,
                width=1.10\textwidth,
                height=0.85\textwidth,
                xlabel={Reward},
                ylabel={Frequency},
                ymin=0,
                ymax=90,
                xmin=-1.0,
                xmax=1.0,
                xtick={-1,-0.75,-0.5,-0.25,0,0.25,0.5,0.75,1.0},
                xticklabel style={rotate=45},
            ]
            \addplot[fill=blue!30, draw=black] coordinates {
                (-1,81) (-0.75,3) (-0.5,7) (-0.25,24) (0.0,47) (0.25,21) (0.5,12) (0.75,5) (1.0,0)
            };
            \end{axis}
        \end{tikzpicture}
    \caption{Before}
    \end{subfigure}
    \hfill
    \begin{subfigure}{0.23\textwidth}
        \begin{tikzpicture}
            \begin{axis}[
                ybar interval,
                width=1.10\textwidth,
                height=0.90\textwidth,
                xlabel={Reward},
                ylabel={Frequency},
                ymin=0,
                ymax=65,
                xmin=0.1,
                xmax=0.9,
                xtick={0.0,0.1,0.2,0.3,0.4,0.5,0.6,0.7,0.8,0.9},
                xticklabel style={rotate=45},
            ]
            \addplot[fill=blue!30, draw=black] coordinates {
                (0.1,3) (0.2,7) (0.3,19) (0.4,33)
                (0.5,52) (0.6,42) (0.7,29) (0.8,15) (0.9,0)
            };
            \end{axis}
        \end{tikzpicture}
    \caption{After}
    \end{subfigure}
\caption{Score distributions before and after RL training.}
\label{fig:histogram}
\end{figure}

\begin{table}[t]
    \small
    \centering
    \setlength{\tabcolsep}{1.5mm}
    \caption{Case study on MSMARCO.}
    \begin{tabular}{l|p{0.35\textwidth}}
        \midrule
        Query     & which health care system provides all citizens or residents with equal access to health care services?  \\
        \midrule
        Groundtruth  & Universal Health Care which is also known as universal care, universal coverage or universal health coverage is a term that is used to address a health care system which provides health care and financial protection to every citizen of a specific country.  \\
        \midrule
        Prediction  & In Singapore all residents receive a catastrophic policy from the government coupled with a health savings account that they use to pay for routine care. In other countries like Ireland and Israel, the government provides a core policy which the majority of the population supplement with private insurance.  \\
        \bottomrule
    \end{tabular}
    \label{tab:case-study}
\end{table}

We evaluate the contribution of RL and present case studies. We also response patterns analysis (Appendix~~\S\ref{sec:resp-pattern}) of \name{}.

\textbf{Effectiveness of RL.}
Figure~\ref{fig:histogram} illustrates the impact of reinforcement learning on model performance.
Before RL training, the model generates outputs with scores spanning -1.0 to 0.75, exhibiting a flatter distribution with an average score of -0.39.
This substantial variance reflects the model has difficulty in stably generating high-quality reasoning paths and embedding tokens.
After RL training, scores concentrate sharply around 0.5, with 69\% of outputs achieving scores above 0.5.
This transformation demonstrates that RL successfully guides the model toward consistently higher-quality outputs, resulting in more deterministic and reliable reasoning and embedding representation.

\textbf{Case Study on MSMARCO.}
We report one scenario we found \name{} demonstrates divergence from the MSMARCO dataset labels, i.e., after enabling reasoning, the retrieval performance degraded.
As illustrated in Table~\ref{tab:case-study}, the query ``which health care system provides all citizens or residents with equal access to health care services?'', the ground truth passage directly answers with ``Universal Health Care'' and provides its definition and scope.
However, our model assigns a higher similarity score to a passage marked as negative, which discusses specific implementations in Singapore, Ireland, and Israel.
Our model tends to prioritize passages that directly answer queries with explicit, concrete examples. In this case, during reasoning, specific keywords like ``Singapore'' are generated and contribute to the embedding representation, as the model recognizes these as essential context to healthcare system queries.
While the ground truth passage delivers a concise definitional answer, our model recognizes the negative passage as more relevant due to its wider contextual coverage.
This observation does not necessarily indicate a quality problem with either the model or the dataset.
Rather, it highlights that this type of search query often requires combining additional contextual signals such as user geographic location or search intent, and typically benefits from a reranking model after the initial retrieval stage.

\section{Related Work}~\label{sec:related}

\subsection{Language Model Adaptation}

Adapting pre-ptrained LLMs to specialized tasks has produced several methodological paradigms, each offering distinct advantages.
Parameter-efficient fine-tuning (PEFT) techniques, such as prefix tuning~\cite{li2021prefix} and LoRA~\cite{hu2022lora}, adjust only a small fraction of model parameters while retaining the model's core functionality.
A particularly relevant approach to our work is instruction tuning~\cite{sanh2021multitask, liu2023vlm, kopf2023openassistant}, which aligns models with tailored tasks or human preferences by training them on specialized examples.
This method has shown substantial improvements in zero-shot generalization, as evidenced by models like FLAN~\cite{longpre2023flan} and Alpaca~\cite{taori2023alpaca}.
Additionally, SGPT~\cite{muennighoff2023sgpt}, INSTRUCTOR~\cite{su2023instructor}, Instruction Embedding~\cite{li2024instructembed} and Qwen3-Embedding~\cite{zhang2025qwen3embedding} explore instruction-tuned embedding representation but simply extract embeddings from model outputs without utilizing the sophisticated reasoning capabilities of LLMs.
RankGPT~\cite{sun2023rankgpt} is a re-ranking model for retrieval tasks, this re-ranking model is different to our embedding models as it takes both the query and multiple candidates as input, and then directly produces a ranking score.
Our method extends this paradigm and demonstrates that instruction tuning can teach LLMs to generate high-quality embeddings through their native token generation process.

\subsection{Augmenting Language Models}

LLMs are known to suffer from fundamental limitations including hallucination, outdated knowledge, and non-transparent reasoning processes, which significantly impact their reliability in knowledge-intensive applications.
To address these challenges, Retrieval-Augmented Generation (RAG) has emerged as a prominent solution that incorporates knowledge from vast and dynamic external databases to enhance the accuracy and credibility of generation~\cite{gao2023rag, yu2024rag, zhao2024rag}.
Existing RAG methods include iterative retrieval that enables multiple retrieval cycles for complex queries, and recursive retrieval that recursively iterates on outputs to process specific task demands~\cite{gao2023rag}.
Other approaches focus on memory enhancement through complementary frameworks such as LongMem~\cite{wang2023langmem}, Camelot~\cite{he2024camelot} and Larimar~\cite{das2024larimar}, which enable LLMs to memorize long history using tailored memory modules.
Current research also focuses on Graph-based RAG~\cite{edge2024graphrag, han2024graphrag} that utilize structured knowledge representation to capture entity relationships and enable multihop reasoning through structure-aware knowledge retrieval.
Our proposed \name{} is orthogonal to all these existing LLM augmentation methods, as it focuses on improving the fundamental representation mechanisms, and integration of our approach with these techniques can bring better system performance through enhanced semantic understanding and more effective model utilization.

\section{Conclusion}

This paper introduces a novel approach \name{}, for transforming LLMs into powerful embedding generators.
Our two-stage approach combines instruction-guided representation learning with reinforcement learning optimization, preserving the model's original architecture and functionality while enabling embedding generation capabilities.
Experiments confirm that \name{} achieves strong performance and validate the effectiveness of our design.
This novel approach represents a substantial advancement in handling complex knowledge-intensive tasks requiring both sophisticated reasoning and effective information retrieval.

\section{Impact Statement}

This paper presents work that advances the field of machine learning by enabling LLMs to perform both reasoning and embedding generation within a unified framework.
To the best of our knowledge, we are the first to enable embedding representation in a chain-of-thought reasoning process. This allows AI-driven applications to use a single unified model for both generative and representative tasks.
The primary societal impact relates to improved computational efficiency and accessibility. By reducing the need for separate models, our approach could lower computational costs and carbon footprint, making advanced AI capabilities more accessible to resource-constrained organizations.


\bibliographystyle{ACM-Reference-Format}
\bibliography{sample-base}

\clearpage
\appendix
\section{Discussion}~\label{sec:discussion}

\subsection{RL Reward Justification}~\label{subsec:disc-reward}

The Stage-2 reward function $\text{DCG}_{\text{scaled}}$ (Equation~\ref{eq:rl-reward}) is designed to satisfy two requirements specific to our RL training regime, and it is the minimal design that satisfies both simultaneously.

\textbf{Continuous reward signal.}
GRPO (Section~\ref{sec:background}) computes per-response advantages by normalizing rewards within a group: subtracting the group mean and dividing by the group standard deviation.
This group-relative normalization is the core mechanism that makes GRPO effective; it requires that responses within a group carry different rewards to produce a meaningful gradient signal.
When all sampled responses in a group receive the same reward, the standard deviation collapses to zero, advantage estimates become undefined, and training stalls.

As the model matures during RL training, the DCG ranking term alone begins to saturate: most of the $G=16$ sampled responses per query achieve near-identical top-rank retrievals (rewards collapse to $[1.0, 1.0, \ldots, 1.0]$).
Any purely rank-based signal, plain nDCG, binary hit/miss, or additive ranking objectives, saturates in the same way, because they all discretize retrieval outcomes.
The reward therefore \emph{must} include a continuous component to remain informative throughout training.

\textbf{The Scale term is the minimal extension.}
Among continuous signals, the cosine similarity between the query embedding and the ground-truth document embedding is the most natural choice:
(1) it is already optimized during Stage-1, requiring no new hyperparameters;
(2) it is bounded in $[-1,+1]$, making it well-scaled for multiplication with the DCG term;
and (3) different reasoning paths produce slightly different embeddings, yielding slightly different cosine similarities, which keeps reward variance non-zero and GRPO gradients alive.

Removing the Scale term can cause training to stall entirely once the model is competent.
Replacing it with a learnable head or an additional reward model would add parameters, hyperparameters, and training complexity.
The multiplicative $\text{Scale} \times \text{DCG}$ form is therefore the minimal, parameter-free extension that prevents training stall while preserving the retrieval-quality interpretation of the reward.

\subsection{Comparison with LLM2Vec}~\label{subsec:disc-llm2vec}

LLM2Vec~\cite{behnamghader2024llm2vec} also adapts decoder-only LLMs for text embedding.
However, LLM2Vec pursues a fundamentally different architectural strategy: it converts a decoder-only LLM into a bidirectional BERT-style encoder by enabling full attention across all token positions and applying masked next-token prediction.
After this conversion, the model no longer functions as an autoregressive LLM: it cannot perform next-token generation, does not support KV-cache reuse, and loses compatibility with standard LLM inference infrastructure such as chunked prefill.

\name{} takes the opposite design choice: it retains the complete decoder-only architecture without modification, extracting embeddings from the hidden state of the special \embedding{} token after autoregressive reasoning.
This preserves all standard LLM capabilities and is orthogonal to inference-time optimizations.

In our evaluation on five public retrieval benchmarks, \name{}-Large achieves overall comparable nDCG@10 scores to LLM2Vec-8B.
The differences are most pronounced on tasks that benefit from reasoning and deeper semantic understanding: on MKQA-eng, \name{}-Large scores $0.352$ compared to $0.226$ for LLM2Vec-8B, a gap we attribute to the reasoning steps performed prior to embedding. Similarly, \name{}-Large outperforms LLM2Vec-8B on DS1000 and MedicalQA.
LLM2Vec-8B is stronger on LitSearch and SciFact, tasks involving dense scientific text where bidirectional context may provide an advantage.
LLM2Vec-1B underperforms the BGE-M3 baseline, consistent with the results reported in their original paper.

Performance differences at the 7B/8B scale are attributable to differences in training data composition and domain coverage rather than architectural superiority.
The key distinction is that \name{} preserves full decoder-only LLM capabilities while achieving competitive embedding quality, offering a clear engineering advantage for deployment scenarios that require both generation and retrieval in a single model.

\subsection{Ablation of Stage-1 Losses}~\label{subsec:disc-losses}

The Stage-1 composite loss (Equation~\ref{eq:stage-1-loss}) consists of four terms: $L_{\text{SFT}}$, $L_{\text{KL}}$, $L_{\text{InfoNCE}}$, and $L_{\text{TripletMargin}}$.
This is the minimal viable design: each term addresses a distinct failure mode, and no term is redundant.

\textbf{$L_{\text{SFT}}$.}
Without $L_{\text{SFT}}$, the model has no gradient signal to produce the \embedding{} token at all.
Stage-2 RL imposes a hard penalty of $-1.0$ for responses missing \embedding{}, so a model that never produces it cannot be trained with RL.
$L_{\text{SFT}}$ is thus a prerequisite for the entire framework.

\textbf{$L_{\text{KL}}$.}
$L_{\text{KL}}$ prevents catastrophic forgetting of the base model's language understanding during Stage-1 contrastive fine-tuning.
This is standard practice in post-training pipelines~\cite{ouyang2022rlhf, deepseek2025r1}: without it, fine-tuning on a narrow contrastive objective can collapse the model's general representations.
Since Stage-2 relies on the model's reasoning ability to produce diverse and informative responses, preserving pre-training knowledge is essential: it is an architectural safeguard, not a tunable design choice.

\textbf{$L_{\text{InfoNCE}}$.}
$L_{\text{InfoNCE}}$ provides a global view of the embedding space by contrasting the query against all documents in the batch simultaneously.
This trains the model to produce embeddings that discriminate across a large, diverse set of negatives, establishing a well-distributed embedding space.

\textbf{$L_{\text{TripletMargin}}$.}
$L_{\text{TripletMargin}}$ adds an explicit cosine distance constraint through curated hard-negative triplets.
Crucially, it directly optimizes the same cosine distance metric used in the Stage-2 RL reward (the $Scale$ term).
Without $L_{\text{TripletMargin}}$, cosine distance is only implicitly constrained through the dot-product similarity in InfoNCE; the margin between positives and hard negatives is not explicitly enforced, and the Stage-2 reward signal operates on a metric that Stage-1 never directly optimized.

In summary, $L_{\text{SFT}}$ enables format compliance, $L_{\text{KL}}$ preserves general language capability, $L_{\text{InfoNCE}}$ shapes the global embedding distribution, and $L_{\text{TripletMargin}}$ enforces the specific distance metric used in RL.
These roles are non-overlapping, and the design is not over-parameterized.

\section{Response Patterns}~\label{sec:resp-pattern}

We examine representative examples on code analysis and definitional queries, identifying consistent patterns that distinguish high-quality responses from ineffective ones, showing the necessity of RL-based optimization.
For simplicity, we omit the final sentence, ``Therefore, the embedding is \embedding{}.''

\subsection{Case 1: Code Explanation}

\begin{figure}[t]
    \centering
    \begin{lstlisting}[
        frame=single,
        breaklines=true,
        basicstyle=\scriptsize
    ]
    def get_mean(arr,x,y):
    if 1 < x <= len(arr) and 1 < y <= len(arr):
       return (sum(arr[:x])/x+sum(arr[-y:])/y)/2
    return -1
    \end{lstlisting}
    \caption{Case 1: the query.}
    \label{fig:case-1-query}
\end{figure}

\begin{figure}[t]
    \centering
    \begin{lstlisting}[
        frame=single,
        breaklines=true,
        basicstyle=\scriptsize
    ]
    1. **Core concepts**: The function `get_mean` calculates the mean of two segments of an array: the first `x` elements and the last `y` elements. Key components are boundary conditions, slicing operations, and mean calculation.
    2. **Key terminology**: 
    - **Mean**: Arithmetic average of numbers
    - **Slicing**: Python syntax `arr[:x]` and `arr[-y:]` to extract subarrays
    - **Validation**: `1 < x <= len(arr)` ensures valid input bounds
    3. **Synonyms and variations**: "average" for "mean," "sublist" for "segment," "valid indices" for "boundary conditions."
    4. **Contextual information**: This function is useful for analyzing array segments where `x` and `y` are known. Applications include statistical summaries, data segmentation, or anomaly detection in time series.
    \end{lstlisting}
    \caption{Case 1: good response (score=0.79).}
    \label{fig:case-1-good}
\end{figure}

\begin{figure}[t]
    \centering
    \begin{lstlisting}[
        frame=single,
        breaklines=true,
        basicstyle=\scriptsize
    ]
    1. First, I identify that the function aims to calculate a custom mean of a subset of array values.
    2. The function checks for valid index ranges before proceeding, using inclusive bounds for subarrays.
    3. The arithmetic mean calculation combines two subsets: the first `x` elements and the last `y` elements.
    4. Division by 2 is implied to average these two subarray means.
    5. Valid indices (x, y) must satisfy the condition `1 < x, y <= len(arr)`, indicating that at least two elements must be considered in each subarray.
    \end{lstlisting}
    \caption{Case 1: bad response (score=0.13).}
    \label{fig:case-1-bad}
\end{figure}

We observe a divergence in how responses approach technical explanation in the code analysis query (Figure~\ref{fig:case-1-query}).
The high-performing response (Figure~\ref{fig:case-1-good}) demonstrates explicit structuring through categorical organization, separating core concepts, terminology definitions, semantic variations, and practical applications into distinct, labeled sections.
Critically, the response contextualizes the code within real-world use cases such as statistical summarization and anomaly detection.
In contrast, the low-performing response (Figure~\ref{fig:case-1-bad}) exhibits sequential, observation-based structure without organizational hierarchy, enumerates code behaviors without establishing conceptual frameworks, and lacks contextualization.
In this case, code explanation requires domain contextualization beyond technical accuracy, and RL addresses this by rewarding responses with better organization, synonym mappings, and application examples, steering \name{} toward structured, contextual explanations.

\subsection{Case 2: Web Query}

\begin{figure}[t]
    \centering
    \begin{lstlisting}[
        frame=single,
        breaklines=true,
        basicstyle=\scriptsize
    ]
    what is an egret
    \end{lstlisting}
    \caption{Case 2: the query.}
    \label{fig:case-2-query}
\end{figure}

\begin{figure}[t]
    \centering
    \begin{lstlisting}[
        frame=single,
        breaklines=true,
        basicstyle=\scriptsize
    ]
    1. **Core concepts**: An egret is a large, wading bird of the heron family (Ardeidae). It is distinct from an albatross, which is a different seabird family (Diomedeidae).
    2. **Key terms**: "Wading bird" (characteristic for its ability to walk on soft or shallow water), "heron" (closely related family), "plumage" (feathers, often white in some species for mating displays), "heronry" (colonial nesting behavior in trees).
    3. **Synonyms/related terms**: Egret can be called "white heron" or "crane bird," while synonyms include "heron." Related terms might include "pelican" or "cormorant," which are also aquatic birds.
    4. **Context**: Common egrets, like the Snowy Egret or Cattle Egret, often live near freshwater wetlands, lakes, or coastal areas, and are frequently found in mixed-species wading bird communities.
    \end{lstlisting}
    \caption{Case 2: good response (score=0.53).}
    \label{fig:case-2-good}
\end{figure}

\begin{figure}[t]
    \centering
    \begin{lstlisting}[
        frame=single,
        breaklines=true,
        basicstyle=\scriptsize
    ]
    1. Identify the core concept: An egret is a wading bird in the heron family (Ardeidae).
    2. Add defining features: Long neck, legs, and beak; typically white plumage or gray/white mottled; slender build with a curved bill.
    3. Contextualize: Often found in wetlands, wet meadows, or marshes; can stand on one leg when waiting for prey; may feed on fish, insects, or small amphibians.
    \end{lstlisting}
    \caption{Case 2: bad response (score=0.03).}
    \label{fig:case-2-bad}
\end{figure}

For definitional queries (Figure~\ref{fig:case-2-query}), the key difference between high and low quality responses lies in semantic construction versus superficial feature enumeration.
The high-quality response (Figure~\ref{fig:case-2-good}) employs two strategies: it distinguishes the query subject from commonly confused alternatives, and it describes how the subject relates to other species in its environment.
The low-scoring response (Figure~\ref{fig:case-2-bad}) adopts a checklist enumeration approach, includes trivial details that add minimal conceptual value, and exhibits redundancy by repeating classification information.
Effective definitions require semantic richness and functional reasoning rather than attribute listing. RL captures these distinctions by rewarding disambiguation, functional explanations, and relational context.

\end{document}